\pdfoutput=1

\documentclass[11pt]{article}

\usepackage{acl2021}
\aclfinalcopy

\usepackage{times}
\usepackage{latexsym}

\usepackage[T1]{fontenc}

\usepackage[utf8]{inputenc}

\usepackage{microtype}
\usepackage{graphicx}
\usepackage{multicol}
\usepackage{amsmath}
\usepackage{colortbl}

\newcommand{\eat}[1]{}

\newcommand{\sys}{DiP}
\newcommand{\peko}{PeKo}

\definecolor{navyblue}{rgb}{0.2, 0.2, 0.6}
\definecolor{rgreen}{rgb}{0.04, 0.4, 0.13}

\newcommand{\nblue}[1]{\textcolor{navyblue}{\textbf{#1}}}
\newcommand{\green}[1]{\textcolor{rgreen}{#1}}
\newcommand{\red}[1]{\textcolor{red}{#1}}

\title{Toward Diverse Precondition Generation}

\author{{\bf Heeyoung Kwon}$^1$, {\bf Nathanael Chambers}$^2$, and {\bf Niranjan Balasubramanian}$^1$ \\
  $^1$Stony Brook University, Stony Brook, New York \\
  $^2$ US Naval Academy, Annapolis, MD \\ 
  \texttt{\{heekwon, niranjan\}@cs.stonybrook.edu}\\
  \texttt{nchamber@usna.edu} \\}

\begin{document}
\maketitle
\begin{abstract}

Language understanding must identify the logical connections between events in a discourse, but core events are often unstated due to their commonsense nature.
This paper fills in these missing events by generating \emph{precondition events}.
Precondition generation can be framed as a sequence-to-sequence problem: given a target event, generate a possible precondition. However, in most real-world scenarios, an event can have \emph{several} preconditions, requiring diverse generation -- a challenge for standard seq2seq approaches.
We propose \sys, a \textbf{Di}verse \textbf{P}recondition generation system that can generate unique and diverse preconditions. \sys\ uses a generative process with three components -- an event sampler, a candidate generator, and a post-processor. The event sampler provides control codes (precondition triggers) which the candidate generator uses to focus its generation. 
Unlike other conditional generation systems, \sys\ automatically generates control codes without training on diverse examples. Analysis against baselines reveals that \sys\ improves the diversity of preconditions significantly while also generating more preconditions.
\end{abstract}
\section{Introduction}

Preconditions are an important part of language understanding with numerous applications, ranging from event understanding to story generation. 
They provide the semantic glue to understand (or generate) the chains of events common in narrative text.
How can we build intelligent systems to fill in these chains, or to identify semantically related events in context?
\citet{kwon2020modeling} took a first step by introducing a precondition generation task, where given a target event mention the goal is to generate text that describes a precondition for the target. They released the `PeKo' dataset for training, and showed that a GPT-2 model can be fine-tuned on input/output sequence pairs. 

While PeKo is useful, it is constrained by annotating a single relation for each target event. This is contrast to the real-world where most events have many preconditions. For example, ``\textit{opening} a door'' has several preconditions like \textit{approaching} the door, \textit{turning} a key in the door, and \textit{pushing} the door.
PeKo's annotation limits the ability of models to learn to generate multiple and \emph{diverse} preconditions\footnote{In order to observe diverse preconditions for the same (or similar) target event we would need a much larger training set.}. In this work, we address the challenge of generating more preconditions for each target event while still maintaining quality.

Generating non-repetitive diverse outputs is a challenge for any conditional language generation system. Our analysis of the GPT-2 based model shows that this is also the case for preconditions. Table~\ref{tab:beam_examples} shows such top preconditions for an example event. Standard sampling techniques produce high-levels of lexical and semantic redundancy. In the absence of any explicit mechanisms to force diversity, the model just produces minor variations of the same event as preconditions. To obtain diverse candidate preconditions, we have to start looking lower in the model's ranked lists of probable preconditions, thereby sacrificing quality. 

\begin{table}[tbp]
    \centering
    \begin{small}
    \begin{tabular}{l}
         TARGET: [BLANK] to {\bf fill} Mr. Lavelle 's seat, for a term\\
         that expires on Dec. 31, 2008.\\\hline\hline
         The Senate {\bf voted} overwhelmingly on Thursday\\
         The Senate {\bf voted} on Wednesday\\
         The Senate {\bf voted} overwhelmingly on Wednesday\\
         The Senate {\bf voted} overwhelmingly on Tuesday\\
         Mr. Lavelle was {\bf appointed} by Gov. Eliot Spitzer\\\hline
    \end{tabular}
    \end{small}
    \caption{Top 5 preconditions generated from GPT-2 with beam search decoding. Key problem: the top 4 preconditions are almost identical.}\vspace{-0.15in}
    \label{tab:beam_examples}
    \vspace{-0.8em}
\end{table}

How can we induce a model to generate diverse outputs without losing quality? Context sensitivity might help with quality, but it also hinders diversity. To address this we introduce a three-stage generative process, which we call \sys. In the first stage, \sys\ uses an event sampler whose only goal is to generate event trigger words as precondition candidates. In the second stage, \sys\ forces the generative model to use the candidate triggers from the first stage to produce the full description of the precondition event. In the third stage, \sys\ re-ranks and filters the generated descriptions using a precondition classifier (also trained from the same training data).\footnote{We will release the source code upon acceptance.}
A brief example is shown here:

    \begin{center}
    \begin{small}
    \begin{tabular}{p{1.5cm}p{2.5cm}p{2cm}}
    \multicolumn{3}{p{6.5cm}}{\textbf{Target Event}: I apologized for the debacle of the day before, and \texttt{[BLANK]} to \textbf{help} me make it right}\\\\
    \quad \textbf{Stage 1} & \textbf{Stage 2} & \textbf{Stage 3}\\
    \quad trying & I am \textbf{trying} now & \emph{delete} \\
    \quad use &  I \textbf{use} my time & \emph{delete} \\
    \quad used & I \textbf{used} my time & \#2 \\
    \quad asked & \textbf{asked} me & \#3 \\
    \quad hired & \textbf{hired} a new staff & \#1 \\
    \end{tabular}
    \end{small}
    \end{center}

Experiments on the \peko\ dataset show that \sys\ produces more diverse and better quality preconditions compared to standard beam decoding, as well as an iterative filtering extension that applies a standard repetition penalty in a sampling strategy. Analyses show that \sys\ is able to better balance the need for diversity against quality. While the iterative repetition penalty method generates lexically diverse outputs, it often introduces irrelevant information rather than producing distinct types of preconditions. Our human evaluation shows that \sys\ on the other hand is able to produce text that is more likely to be preconditions.

All code and data are available at \url{https://stonybrooknlp.github.io/DiP/}. 


\section{Related Work}

Most work on logical preconditions has focused on identification/extraction from text. For example, 
\citet{sil2010extracting} identified preconditions using a SVM-based score function with hand-crafted PMI and WordNet based features. \citet{branavan2012learning} extracted domain-specific precondition relations from instructions for the game of Minecraft. This paper is instead focused on generating novel preconditions. To the best of our knowledge, only the prior PeKo work~\cite{kwon2020modeling} has attempted this. We are building on those initial ideas.

There has been research for diverse generation using control codes or latent variables. Some works use explicit cues to control text generation. \citet{huang-etal-2018-automatic} used emotion embeddings to generate dialogue responses in a specific mood. \citet{keskar2019ctrl} trained a LM with human readable control codes, which describe domain, style, or topics. Then the model learns to generate text conditioned on a given code. The model requires manually predefined control codes and a corresponding training corpus for each code.

Other diverse generation works learn latent representations or codes from input text, and then generate text conditioned on those codes. \citet{shu-etal-2019-generating} applied a sentence embedding to generate syntactically diverse translations. They find that syntax-based encoding with TreeLSTM~\cite{socher2011dynamic} yields better diversity than a contextual encoding 
using BERT~\cite{devlin-etal-2019-bert} or FastText~\cite{bojanowski2017enriching}.
\citet{bao-etal-2020-plato} used $K$ categorical latent variables to control the generation context of dialogue responses and pick the highest probability response from the responses generated using the latent variables.
COD3S~\cite{weir-etal-2020-cod3s} is designed to generate diverse causal relations. It uses locality-sensitive hashing (LSH)~\cite{indyk1998approximate} on representations from Sentence-BERT~\cite{reimers-gurevych-2019-sentence}. Conditioning on these 16-bit LSH signatures, it generates cause/effect sentences using a Transformer architecture~\cite{vaswani2017attention} but with a limited vocabulary size of 10K. 

These previous approaches have some drawbacks -- they either require explicit control codes and training examples, or they have low interpretability of their codes. Our approach addresses these two limitations: control codes are learned from non-diverse input text and the codes are human-readable events. And these approaches are not directly comparable to our method without proper modification, which would not be fair comparisons. Thus, we present our own baselines for evaluation, and these baselines serve as a proxy of ablation studies as well.
\section{Diverse Precondition Generation} \label{sec:gen}
This section describes our diverse precondition generation task and our methodology for solving it. Our proposed approach does not require additional diverse training examples.

\subsection{Generation Task} \label{sec:gen_task}

This paper follows the precondition definitions from \citet{kwon2020modeling}: 

\noindent\textbf{Precondition Definition} -- "Given a target event mention \textit{t} and a candidate event mention \textit{p}, we assert \textit{p} is \textit{a precondition event for t} if \textit{p} is necessary for \textit{t} to happen i.e., \textit{t} likely would not have occurred without \textit{p}, in the current text context."

\noindent\textbf{Precondition Generation} -- "Given a target event \textit{t}, generate an event \textit{p} that is a precondition for \textit{t}."

The precondition generation task is defined over sentences that contain both a target and a precondition event. The precondition part is masked and a model is asked to reconstruct the sentence including its precondition.
For masking, the syntactic subtree of a precondition is replaced with \texttt{[BLANK]}. In order to indicate the events of interest -- target and precondition -- we use special tokens \texttt{<event>} .. \texttt{</event>} and \texttt{<pre>} .. \texttt{</pre>}. 

For our new task, instead of generating the entire sentence, we only generate a precondition clause that would fit into the input's \texttt{[BLANK]}. Since a precondition could be stated in either preceding or succeeding position of its target event, we modeled this as a text infilling task. This approach is inspired by \citet{donahue-etal-2020-enabling} and this modification allows the model to focus solely on generating preconditions because the model doesn't need to copy over its input text. Thus, the model can learn faster and more efficiently. 

\begin{figure}[ht!]
    \centering
    \includegraphics[width=0.27\textwidth]{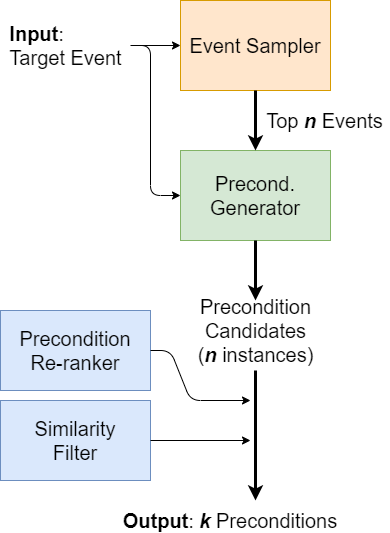} 
    \caption{The \sys\ pipeline. Candidates are generated conditioned on the Event Sampler. The Re-ranker and Similarity Filter improve quality/diversity.} \vspace{-0.15in}
    \label{fig:pipeline}
\end{figure}


\subsection{Diverse Precondition Generator} \label{sec:dip}

Generating preconditions is a difficult task even for a single output setting~\cite{kwon2020modeling}.
With the training data derived from existing news articles, generative models only get to see one possible precondition for each target event. Not surprisingly the top candidates in beam search tend to be focused towards a specific type of precondition event with minor variations. This suggests that we need to provide explicit guidance to the model to explore diverse candidates.

How can we get such diverse guidance? A main strength of large generative language models is that they learn to generate text that fits with the input context. If we can get the input context to be less specific then we can aim to get more general outputs. We can exploit this behavior by training a separate event sampler that is fed a reduced version of the target event description. For example, we can denote the target event by just the event trigger and its arguments. The event sampler learns to predict possible precondition event triggers based on this reduced context. This task forces the sampler to learn a more general mapping between target and precondition events that can produce a diverse set of starting points for generating the precondition events. We can then train another generative model to condition on the precondition trigger in addition to the input sentence. This gives us a model whose outputs we can control by providing different possible precondition triggers. Not all precondition triggers may yield high quality preconditions. To further assist the model, we also devise a precondition re-ranker.

Our overall system, shown in Figure \ref{fig:pipeline}, consists of three components -- an event sampler, a candidate generator, and a post processor (Precondition re-ranker and Similarity filter). The first two stages are used for generation -- they use two separate generation models, and the last is employed to improve the quality of generated preconditions. We refer to this system as \sys\, short for Diverse Preconditions.

\subsubsection{Event Sampler}

The event sampler provides possible precondition event triggers given a target event. This can be formulated as a sequence to sequence problem where the input sequence is a target event and the output sequence is a precondition event trigger. Since our goal here is to get diverse precondition events, we can experiment with input contexts of different levels of detail. To get more general precondition events, we use just the target event triggers as the input. To get more specific preconditions, we can use larger contexts surrounding the target event trigger as the input. During inference, we sample top $n$ event triggers based on their probability.

Formally, let $x'$ be a subset of the full description $x$ of the target event. The sampler can be seen as a generative model that outputs event triggers $e$ for the preconditions of the target event.
$$ \hat{e} = \arg\max_e \log p(e|x')$$

The generative model is trained to maximize the probability for the correct precondition trigger $e$ and during inference can be used to sample a candidate set of top $n$ precondition event triggers.

\subsubsection{Candidate Generator} 

The candidate generator, as the name suggests, is a language model that we fine-tune for generating precondition candidates. We want this model to generate preconditions corresponding to the triggers from the event sampler. To this end, in addition to the full target event description $x$, we also provide a precondition trigger marked by a special token -- \texttt{<E> \textit{precondition\_event}} -- at the end of the input. This can be seen as a form of a control code similar to those used in \citet{keskar2019ctrl, weir-etal-2020-cod3s}. The crucial difference, however, is that the codes in our case are dynamically generated conditioned on the input and not restricted to a predefined set.

Formally, the candidate generator is a language model that generates a description of the precondition event $c_i$ conditioning on the full description of the target event $x$ and a given precondition trigger $e_i$ from the event sampler.
$$ \hat{c_i} = \arg\max_{c_i} \log p(c_i|x,e_i)$$
%

The model is trained to maximize the probability of the observed precondition text for the target event when provided with the correct precondition trigger. 
Note that during training, the precondition trigger provided as input always appears in the correct precondition description output ($\hat{c_i}$). This encourages the model to learn to incorporate the trigger provided at the end of the input as part of its output. During inference, the model generates a set of preconditions one for each of the top $n$ triggers obtained from the event sampler.  

\subsubsection{Post Processor}

\noindent{\bf Precondition Re-ranker} We use a precondition re-ranker to reorder the generated candidates based on how likely they are to be preconditions of the target event. Note that the generative model is implicitly trained for a similar objective. However, the model is also forced to include the input precondition trigger which could make it harder to focus on ensuring the result is indeed a precondition. Therefore, we introduce a separate precondition classifier that scores the generated candidates. Note that the original \peko\ dataset is already setup for training such a classifier~\cite{kwon2020modeling}. Each instance in this dataset consists of an input text that includes a pair of marked event triggers (target, candidate) and a label that indicates whether the candidate is a precondition of the event denoted by the target trigger. The output from the precondition generator is essentially equivalent to an instance from this dataset. We build a classifier that scores a pair of events in text, and we use this score as an indicator of the precondition quality of the generated candidates and re-rank them based on this score.

\noindent{\bf Iterative Redundancy Filtering} The resulting candidates are a mix of candidate precondition events from different triggers. To further avoid redundancy we also include an explicit filtering step, where we post-process the generated text based on their pairwise similarity. Specifically, we start with the highest ranked instance in the output set, and iteratively walk down the ranked list and add instances to the output if the highest similarity score they have with any of the current output set is lower than a certain threshold. 

\section{Evaluation}

\begin{table*}[tbh!]
    \centering
    \begin{small}
    \begin{tabular}{p{2cm}|p{13cm}} \hline\hline
    Target Event & The Metropolitan Transportation Authority recently \textit{\textbf{canceled}} some large projects \\ \hline
    Trigger Only & took, canceled, succeeded, began, scheduled, died, rejected, decision, filed, pushed \\
    $\pm3$ tokens & planned, took, began, expected, needed, approved, designed, intended, completed, brought\\
    $\pm5$ tokens & planned, intended, took, began, aimed, devised, completed, expected, designed, needed \\ \hline\hline
    Target Event & They tried to \textit{\textbf{rebuild}} their shattered nation \\ \hline
    Trigger Only & took, rebuilt, losing, sustained, lost, opened, came, bought, died, used \\
    $\pm3$ tokens & took, lost, war, losing, reached, brought, moved, began, fled, came \\
    $\pm5$ tokens & took, lost, war, collapsed, left, failed, abandoned, came, laid, began \\ \hline
    \end{tabular}
    \end{small}
    \caption{Top 10 generated event triggers from the event sampler. As more context is provided, the model generate more specific events related to the provided context.}
    \label{tab:sampling_examples}
    \vspace{-0.5em}
\end{table*}

\begin{table*}[tbh!]
    \centering
    \begin{small}
    \begin{tabular}{p{10cm}|p{5cm}}
    Input Text & Generation Target \\\hline\hline
     \texttt{[BLANK]} that donations be \textit{\textbf{made}} to the Crohn's and Colitis Foundation or NYBOT Futures and Options for Kids in memory of Harry. \texttt{<E>} requests & In lieu of flowers, the family \textit{\textbf{requests}} \\ \hline
     \texttt{[BLANK]} to \textit{\textbf{start}} trading an important Nymex product, West Texas intermediate crude oil. \texttt{<E>} inspired & Nymex's foray also \textit{\textbf{inspired}} ICE \\ \hline
      Mr. Robbins played hard and fluidly, \texttt{[BLANK]} to \textit{\textbf{give}} his solos funk and shape. \texttt{<E>} landing & \textit{\textbf{landing}} heavily on unexpected notes \\\hline
    \end{tabular}
    \end{small}
    \caption{Examples of training instance pairs for the candidate generator. Unlike \citet{kwon2020modeling}, we add the precondition event at the end of the input to help the model utilize the event trigger when generating a precondition.}\vspace{-0.15in}
    \label{tab:training_examples}
    \vspace{-0.5em}
\end{table*}

Our goal is to investigate the impact of our \sys\ approach for generating diverse and high-quality preconditions. We closely follow ~\citet{kwon2020modeling} for the experimental setup and the GPT-2 based generation system for our evaluation. 

\subsection{Datasets} \label{sec:dataset}
For the fine-tuning task, we use the precondition generation instances in the \peko\ dataset. In addition, we also create a large additional pre-training dataset that includes temporal generation instances. With this additional dataset we can perform a form of domain adaptive pre-training (DAPT) introduced by \citet{gururangan-etal-2020-dont}. The main idea here is to create generation instances where the model gets to see a target event but now is required to produce an event that temporally precedes the target event. Since preconditions are supposed to be temporally preceding this temporal generation task can be seen as a more permissive yet related generation task, which is then subsequently restricted to only preconditions in the fine-tuning stage. We use the CAEVO~\cite{chambers2014dense} system to obtain temporally related event pairs from the NYT corpus~\cite{nytcorpus}. This yields 1.1 million instances and each instance contains one temporal relation (BEFORE/AFTER). Note that all systems are trained using the same pre-training and fine-tuning strategy using both datasets.


\subsection{Baselines}

\noindent{\bf Beam Search} As a baseline, we use text infilling GPT-2 system (inspired by \cite{donahue-etal-2020-enabling} with a standard beam search decoding strategy. This beam search decoder can provide multiple responses up to its beam size. We expect this simple baseline to contain high-levels of redundancy in its outputs. 

\noindent{\bf Repetition Penalized Sampling (RPS)} For a stronger baseline, we use a decoding strategy that can generate diverse preconditions by penalizing previously generated precondition event triggers. This is done by an iterative decoding process applied to the same GPT-2 generation model. Given a target event, the model generates $k$ preconditions in an iterative manner. When the model generates a precondition trigger -- after \texttt{<pre>} token -- a repetition penalty is applied to deter the model from selecting previously generated precondition events. We adopt the penalized sampling from \citet{keskar2019ctrl}. Instead of using a list of all generated tokens, we use a list of precondition event triggers that are generated in the previous iterations.  Given a list of generated precondition events $t$, the probability distribution $p_i$ for the next trigger token $x_i$ is defined as:
\begin{equation*}
\begin{split}
p_i = \frac{\exp{(x_i/I(i\in t))}}{\sum_j{\exp(x_j/I(j\in t))}} \\
I(c)=\lambda\quad\textrm{if $c$ is true else 1} 
\end{split}
\end{equation*}

We set $\lambda = 1.2$ as in \citet{keskar2019ctrl}.
For decoding, we use Nucleus Sampling~\cite{Holtzman2020The} which has been claimed to generate a higher quality of text.
Finally, we test the RPS model with the post-processor from \sys, to confirm that the major gain of \sys\ is from the Event Sampler.

\subsection{\sys\ Model}

\sys\ has three modules -- Event sampler, Candidate generator, and Precondition re-ranker. We train each module separately.\\

\noindent{\bf Event sampler} We use the GPT-2 model for the event sampler. The model is trained on the same data instances described in Section \ref{sec:dataset}, but instead of using the entire target-precondition pairs, we use target-precondition \textit{event trigger} pairs. We train three event samplers with different levels of context -- trigger only, 3 neighboring tokens, and 5 neighboring tokens -- to understand how different context affect candidate precondition sampling. As Table \ref{tab:sampling_examples} shows, adding more context help the model to generate more specific events related to describe situations while the model provides more general events if only a trigger is given.

\noindent{\bf Candidate generator} The GPT-2 model is also used for the candidate generator. For training, as described in \ref{sec:dip}, we add \texttt{<E> precondition\_event} at the end of input so that the model can learn how to utilize the provided event trigger as a control code. Table \ref{tab:training_examples} shows the training examples for the candidate generator. 

\noindent{\bf Post-processor} 
We train a precondition re-ranker using BERT~\cite{devlin-etal-2019-bert}. The F1 score of the classifier is 71.91 with 64.65 of the precision. To remove possibly redundant preconditions using iterative redundancy filtering, we need to compute cosine similarity between the generated preconditions. We take the precondition classifier's [CLS] token representation as the embedding for preconditions. Since the similarity score distributions are different from instance to instance, instead of using a fixed value as the threshold, we set the threshold as $\mu + \sigma$ of each instance (the mean and the standard deviation of pairwise similarity scores). This filters out $\sim$16\% of the most similar generated preconditions. 
For comparison with the baselines, we take top 10 preconditions from remaining outputs.


\subsection{Automatic Evaluation Metrics} 
We use Self-BLEU~\cite{zhu2018texygen} and Self-BLEURT score to measure the diversity of generated preconditions. Self-BLEU measures how similar a set of sentences is to each other using BLEU score -- the average of BLEU scores for the all pairs of sentences in the set. 
In addition to direct lexical overlap, we also measure semantic overlap using BLEURT~\cite{sellam-etal-2020-bleurt}, which is a BERT-based learned evaluation metric that is trained on human ratings of sentence pairs. We refer to this metric as Self-BLEURT. For both metrics, a lower score implies more diverse preconditions. 

\subsection{Results}
We compare the models on both diversity and quality. For diversity, we use an automatic evaluation, and for quality we used human annotators.\\\vspace{-2mm}

\begin{table}[tbp]
    \centering
    \nblue{Model Diversity Evaluation}\\\vspace{.1in}
    \begin{tabular}{l|r|r}
         \nblue{Model} & \nblue{Self-BLEU} & \nblue{Self-BLEURT}\\\hline\hline
         Beam Search & 0.234 & -0.450 \\
         RPS & 0.016 & -1.273\\
         RPS+Post-proc. & 0.013 & -1.280\\
         \sys & 0.038 & -1.111 \\\hline
    \end{tabular}
    \caption{Diversity evaluation for different models. We evaluate top 10 preconditions for each model. RPS+Post-proc. produces the most diverse outputs followed by RPS and \sys\ with a small margin.}\vspace{-0.15in}
    \label{tab:diversity_model}
\end{table}

\noindent{\bf Automatic Diversity Evaluation}: 

\noindent
Table \ref{tab:diversity_model} shows the diversity metrics for all methods. We evaluated 5,000 preconditions generated for 500 target events. 
Comparing RPS+Post-proc to RPS, Post-proc shows little effect, we compare just RPS to \sys\ in the rest of the evaluations (See Appendix for more details between RPS and RPS+Post-proc).

\begin{figure}[tb!]
    \centering
    \includegraphics[width=0.45\textwidth]{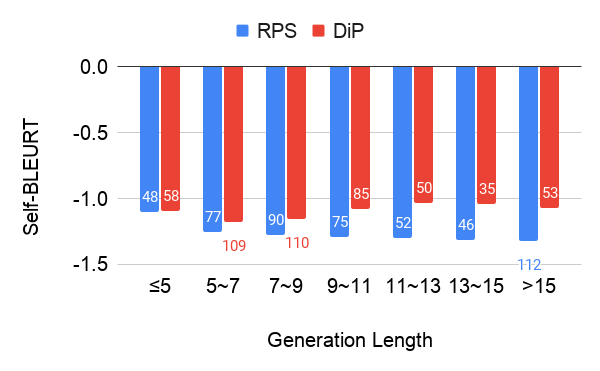} 
    \vspace{-0.2in}
    \caption{The Self-BLEURT scores across different lengths of generated text. The numbers indicate the number of instances in each bucket.
    }
    \label{fig:self_bleurt}
\end{figure}

In both metrics, DiP and RPS generate more diverse output than the beam search decoder. \sys\ is compatible to RPS in shorter preconditions, and RPS produces more diverse outputs when the generated text gets longer, as shown in Figure \ref{fig:self_bleurt}.\\\vspace{-2mm}

\noindent{\bf Manual Quality Evaluation}:

\begin{table}[tb!]
    \centering
    \nblue{Quality of Preconditions}\\\vspace{.1cm}
    \begin{tabular}{l|c|c}
         \nblue{Model} & \nblue{Average Score} & \nblue{\#Wins}\\\hline\hline
         RPS & 0.954 & 30\\
         \sys & \textbf{1.101} & \textbf{56} \\\hline
    \end{tabular}
    \caption{The Top 10 generated preconditions for each target event were scored on a 0-2 scale. A model "wins" a target if its average is highest. Using Bootstrapping with $n=1000$ the 95\% conf-interval for the RPS mean is (0.89, 1.01) and DiP is (1.05, 1.15). 
    }\vspace{-0.1in}
    \label{tab:human_eval}
\end{table}

\begin{figure}[tb!]
    \centering
    \includegraphics[width=0.45\textwidth]{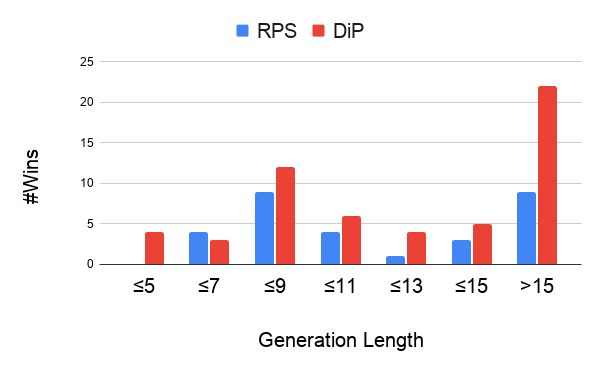} 
    \vspace{-0.2in}
    \caption{The number of wins across generation lengths. \sys\ wins more as the generations lengthen.
    }\vspace{-0.2in}
    \label{fig:win_length}
\end{figure}

\noindent
The automatic evaluation only measures diversity. To see if the models generate legitimate preconditions, we conducted a manual evaluation for quality. We evaluated 960 generated outputs covering 96 distinct target events for both \sys\ and the RPS baseline. For each instance the annotators were presented with the top ten generated outputs from two systems. 
For each output the annotators provided a rating on a scale from 0 to 2, where 0 means not a precondition, 1 is a maybe, and 2 is definitely a precondition. We split the 96 instances across 8 different annotators\footnote{These were computer science graduate students}. 

Table~\ref{tab:human_eval} shows the results in terms of two metrics: one is the average score across all 96 instances, and another is the number of "Wins" where a model gets +1 point if the sum of its 10 precondition scores is higher than the other. In both metrics, \sys\ outperforms RPS. Moreover, as shown in Figure \ref{fig:win_length}, \sys\ produces better preconditions across most output lengths and is best on longer outputs.



\subsection{Analysis}

\noindent{\bf Examples:} Table \ref{tab:gen_examples} shows the top 5 generations from our main three systems. The beam search's failure on the diversity metric is easy to see with its repetitive output. Most verbs are the same. Both RPS and \sys\ are notably better in terms of diversity, but RPS introduces lots of irrelevant information that may have artificially increased its diversity score. Long irrelevant phrases are clear to see, and verb synonyms are common. In contrast, \sys\ generates more succinct and general preconditions, as well as fewer direct synonyms. \\\vspace{-2mm}

\noindent{\bf Context Specificity: }
Table \ref{tab:diversity_context} shows the diversity scores when different levels of context are provided to the event sampler. Diversity gets slightly worse with \emph{more} context. This aligns with our observation from Table \ref{tab:sampling_examples} that the event sampler with more context generates more context-specific precondition events, which now appear to be closer to each other semantically. This makes intuitive sense if you view context as closing a model's view of broader options.\\\vspace{-2mm}

\noindent{\bf Errors: }
We categorize 4 types of frequently observed errors from \sys. 1) \textbf{Rare target events}: some target event triggers are polysemious, and some meanings are rarely used in the newswire. Thus, the event sampler is biased toward generating precondition events based on more frequent usages of triggers. 2) \textbf{Conflict with sentence structure}: the candidate generator has no ability to modify the trigger for its syntactic context because it is trained only as copying over the event trigger. 3) \textbf{Re-ranking issue}: the re-ranker can induce errors due to its performance -- 64.65 precision. 4) \textbf{Limited available context}: when the provided context of a target event is too limited, the model often fails to generate preconditions. Table \ref{tab:error_examples} shows examples for each type of error.

\begin{table}[tb!]
    \centering
    \begin{small}
    \begin{tabular}{|p{1.5cm}|p{5cm}|} \hline
        \multicolumn{2}{|p{6.5cm}|}{\bf Rare target events}\\\hline
        \multicolumn{2}{|p{6.5cm}|}{\texttt{[BLANK]} my homework,'' and exercising like a demon at a nearby gym. She'd \textit{\textbf{lost}} weight and viewed herself as training for a fight.} \\ \hline
        Event & Generated Precondition \\ \hline
        war & I \textit{\textbf{war}} with the other women in my school, who had been working together on \\ \hline
        died & She \textit{\textbf{died}} of cancer in 1998, after her husband was killed by an electrician who had been working with him''\\\hline
        \multicolumn{2}{|p{6.5cm}|}{The word \textit{\textbf{lost}} is usually used in the context of war, election, or obituary in the newswire. That may lead the event sampler to fail to generate proper precondition event triggers.} \\ \hline\hline
        \multicolumn{2}{|p{6.5cm}|}{\bf Conflict with sentence structure}\\\hline
        \multicolumn{2}{|p{6.5cm}|}{The proposal from Mr. Levy that the Justice Department \textit{\textbf{struck}} down, she added, would have violated nondiscrimination provisions of the consent decree and the Civil Rights Act of 1964 by \texttt{[BLANK]}.} \\ \hline
        Event & Generated Precondition \\\hline
        created & \textit{\textbf{created}} a new state law requiring that a judge be present on the court \\ \hline
        called & \textit{\textbf{called}} for a ''state level ban on political activity'' in New York City\\\hline
        \multicolumn{2}{|p{6.5cm}|}{The candidate generator is not able to modify the provided triggers according to the context.} \\ \hline\hline
        \multicolumn{2}{|p{6.5cm}|}{\bf Re-ranking issue}\\\hline
        \multicolumn{2}{|p{6.5cm}|}{China's markets nearly \textit{\textbf{disintegrated}} in 2005, and one 2003 poll found     \texttt{[BLANK]}.} \\\hline
        Event & Generated Precondition \\ \hline
        became (\#4) & that the market \textit{\textbf{became}} more popular than in 2000 (\#2) \\ \hline
        named (>\#10) & that the country was \textit{\textbf{named}} the world's largest economy (\#4)\\\hline
        \multicolumn{2}{|p{6.5cm}|}{Generated sentences are not preconditions but ranked high after re-ranking (\#4 $\rightarrow$ \#2 and >\#10 $\rightarrow$ \#4).}\\ \hline\hline
        \multicolumn{2}{|p{6.5cm}|}{\bf Limited available context}\\\hline
        \multicolumn{2}{|p{6.5cm}|}{\texttt{[BLANK]} to \textit{\textbf{hit}} a short forehand and guided it wide.} \\ \hline
        Event & Generated Precondition \\ \hline
        batted & In the third round, he \textit{\textbf{batted}} the ball with his left hand \\ \hline
        pitched & In the first inning, the Mets \textit{\textbf{pitched}} three consecutive hits\\ \hline
        \multicolumn{2}{|p{6.5cm}|}{The target context is related to tennis, but preconditions are generated in the context of baseball because the provided context is too limited.}\\ \hline
    \end{tabular}
    \end{small}
    \caption{Examples from each type of errors. There are 4 types of frequently observed errors and the first 3 types are caused by each stage in \sys. The last one is due to data instances.} \vspace{-1em}
    \label{tab:error_examples}
\end{table}


\begin{table}[tb!]
    \centering
    \begin{tabular}{l|r|r}
         \nblue{Context} & \nblue{Self-BLEU} & \nblue{Self-BLEURT}\\\hline\hline
         Trigger only & 0.038 & -1.111 \\
         $\pm3$ tokens & 0.039 & -1.103\\
         $\pm5$ tokens & 0.040 & -1.098 \\\hline
    \end{tabular}
    \caption{Diversity evaluation for different samplers. Precondition candidates are generated from the event samplers given the input with a trigger only, a trigger with neighboring 3 tokens, or a trigger with 5 tokens.}
    \label{tab:diversity_context}
\end{table} 

\begin{table*}[tb!]
    \centering
    \begin{small}
    \begin{tabular}{p{4.9cm}|p{4.9cm}|p{4.9cm}} \hline
    \multicolumn{3}{l}{\textbf{Target Event}: \texttt{[BLANK]} to \textit{\textbf{maintain}} below-market rents.} \\ \hline
    Beam Search & RPS & \sys \\\hline
\cellcolor{red!15}In the last few years, real estate prices have \red{\textit{\textbf{risen}}} and real estate prices have dropped &
\cellcolor{red!15}Wilhelmina's stock \red{\textit{\textbf{dropped}}} 8 percent in early 2005 after a lull, but more actively revived the strong dollar, lending to an expected influx of funds & The city has \green{\textit{\textbf{reached}}} a deal with the tenants \\\hline

\cellcolor{red!15}In the last few years, real estate prices have \red{\textit{\textbf{risen}}} and real estate investment trusts have grown & The City Council's 10-member City Planning Committee \green{\textit{\textbf{voted}}} 11 to 6 yesterday & The city will \green{\textit{\textbf{use}}} the money \\\hline

\cellcolor{red!15}In the last few years, real estate prices have \red{\textit{\textbf{risen}}} and real estate prices have dropped, but rents have continued & 
\cellcolor{red!15}property values are \red{\textit{\textbf{rising}}}, rising as the vacancy rate is expected & In the 1980s, the city \green{\textit{\textbf{moved}}} its building to a new site in the East River \\\hline

\cellcolor{red!15}In the last few years, real estate prices have \red{\textit{\textbf{risen}}} and real estate investment trusts have grown in size & In less than a year, such improvements have \green{\textit{\textbf{increased}}} through acquisitions and capitalization at New York City police stations, legal firms and cruise ships, suggesting that housing can be bought & The City Council \green{\textit{\textbf{passed}}} a bill on Wednesday that would give the city the authority to build a new building at the site of the old Erez subway station, and \\\hline

\cellcolor{red!15}In the last few years, real estate prices have \red{\textit{\textbf{risen}}} and real estate prices have dropped, and rents have risen in the last few years & 
\cellcolor{red!15}Matthew Hallico, president of the General Electric Company in Manhattan, and Robert Chrisin, a sales vice partner at Ira G. Albrecht His comments about the incentive package \red{\textit{\textbf{raised}}} many concerns about how it works, as well as what shareholders might do & In the 1980s, the city \green{\textit{\textbf{began}}} a program \\\hline\hline

\multicolumn{3}{p{15cm}}{\textbf{Target Event}: By about 10 p.m., the proposals \textit{\textbf{appeared}} dead for now \texttt{[BLANK]}.} \\ \hline
Beam Search & RPS & \sys \\\hline
Mr. Spitzer \green{\textit{\textbf{took}}} office &
after the judge, Col Richard Kultura of Thailand, \green{\textit{\textbf{signaled}}} the end of his sentence & after the commission \green{\textit{\textbf{filed}}} a proposal to provide \$ 2 million in new money for the project \\ \hline

\cellcolor{black!15}Mr. Spitzer \green{\textit{\textbf{took}}} over & 
\cellcolor{red!15}after city Hall \red{\textit{\textbf{learned}}} it would begin public comment on ways it could add 27,000 new jobs to the island & 
after the Senate 's Democratic majority has \green{\textit{\textbf{taken}}} over control of the House \\ \hline

the City Council \green{\textit{\textbf{voted}}} on them &
as they were \green{\textit{\textbf{rejected}}} by legislative leaders & 
after the State Legislature \green{\textit{\textbf{put}}} them on a vote \\ \hline

\cellcolor{black!15}Mr. Spitzer \green{\textit{\textbf{took}}} office in January &
\cellcolor{red!15}when Mayor Mark Meehan \red{\textit{\textbf{heeded}}} all of his smaller complaints & 
when the State Legislature \green{\textit{\textbf{used}}} the budget to cut a \$ 2 billion tax break \\ \hline

\cellcolor{red!15}the City Council \red{\textit{\textbf{passed}}} them to the City Council &
after the State Senate \green{\textit{\textbf{voted}}} yes on key issues & 
after a suicide bomber \green{\textit{\textbf{killed}}} a man in an Internet chat room\\ \hline\hline

\multicolumn{3}{p{15cm}}{\textbf{Target Event}: \texttt{[BLANK]} to \textit{\textbf{scout}} potential recruits.} \\ \hline
Beam Search & RPS & \sys \\\hline
The N.F.L. and the N.B.A. have \green{\textit{\textbf{taken}}} steps & 
The pending replacement of Carl Crawford has \green{\textit{\textbf{enticed}}} some intelligence officials and top Qaeda leaders &
The police \green{\textit{\textbf{took}}} over the department 's operations , and they began
\\\hline
\cellcolor{black!15}The N.F.L. and the N.F.L. have \green{\textit{\textbf{taken}}} steps &
Employees are \green{\textit{\textbf{giving}}} them the opportunity &
The department is \green{\textit{\textbf{sending}}} a new system
\\\hline
\cellcolor{black!15}The N.C.A.A. has \green{\textit{\textbf{taken}}} steps &
Most Somalis \green{\textit{\textbf{want}}} a law that would enable them &
The New York State Department of Education \green{\textit{\textbf{began}}} a program last year
\\\hline
\cellcolor{black!15}The N.F.L. and the N.B.A. have \green{\textit{\textbf{taken}}} similar steps & 
Shortly after Katrina , Post servicemen were \green{\textit{\textbf{chasing}}} selectors after the storm 's onset & 
The department has \green{\textit{\textbf{sent}}} a handful of officers to the police 
\\\hline
\cellcolor{black!15}The N.F.L. and the N.F.L. have \green{\textit{\textbf{taken}}} similar steps &
\cellcolor{red!15}Ever since college opened in 1983 , he \red{\textit{\textbf{shopped}}} for school assignments & The N.C.A.A. \green{\textit{\textbf{set}}} up a task force\\ \hline
    \end{tabular}
    \end{small}
    \caption{Top 5 generations from 3 systems. Red cells are invalid preconditions. Greyed out cells are repetitions from previous cells. \sys\ produces both valid \emph{and} diverse preconditions.
    }
    \label{tab:gen_examples} \vspace{-0.15in}
\end{table*}
\section{Conclusion} 

Real-world events often have multiple preconditions, but today's datasets do not, including the latest PeKo, presenting a challenge for text-driven models. 
Vanilla generative models have high-levels of redundancy in their outputs and are thus not well suited for diverse generation.
This work introduced an event sampler that overcomes the issue of target context specificity to provide diverse guidance to the generator. 
Coupled with a precondition ranker and similarity filter, this multi-stage generation setup yields more diverse and higher quality preconditions.
Further, a new training corpus was not required.
More generally, this approach can be seen as an instance of controllable diverse output generation for conditional language models.

\section*{Acknowledgements}

This material is based on research that is supported in part by the Air Force Research Laboratory (AFRL), DARPA, for the KAIROS program under agreement number FA8750-19-2-1003 and in part by the National Science Foundation under the award IIS \#2007290.

\bibliography{acl2021}
\bibliographystyle{acl_natbib}

\newpage
\appendix

\section{Appendix}
\label{sec:appendix}
\subsection{Experimental Details}

\subsubsection{Data Split}

For the dataset for pre-training, we split into train/dev/test with the ratio of 8:1:1. For PeKo dataset, we follow the setting from \citet{kwon2020modeling}.

\subsubsection{Infrastructure}

All models are trained using NVIDIA Titan RTX (24GB of GDDR6 VRAM).

\subsubsection{Parameters}

We use \citet{wolf-etal-2020-transformers} library for all transformer models.
For the beam search baseline and RPS model we use the GPT-2 architecture, which has 124,445,184 trainable parameters. DiP model consists of two GPT-2 models for the event sampler and the candidate generator -- 2 $\times$ 124,445,184 -- and one BERT model for the re-ranker -- 108,313,346 parameters.
In total, DiP has 357,203,714 trainable parameters.

\noindent{\bf Optimizer}: We use AdamW~\cite{loshchilov2018decoupled} for the optimizer across all models. For pre-trianing, we fix the learning rate as 1e-3. For fine-tuning, we experiment with [1e-4, 1e-5, 1e-6].

\noindent{\bf Event sampler}: For pre-training, the epochs are set to 100 with the batch size of 128 for the trigger only, 64 for the $\pm3$ tokens, and 32 for the $\pm5$ tokens model. For fine-tuning, the epochs are set to 10 with the batch size of 32. 

\noindent{\bf Candidate generator}: For pre-training, the epochs are set to 100 with the batch size of 16. For fine-tuning, the epochs are set to 10 with the batch size of 16.

\noindent{\bf Precondition re-ranker}: We use the classifier provided by the authors of \citet{kwon2020modeling} -- \url{https://stonybrooknlp.github.io/PeKo/}.

All Models are picked based on the losses from the dev set.

\subsection{Comparison between RPS and RPS+Post-processing}
Table \ref{tab:gen_examples_rps+} shows the comparison between RPS and RPS+Post-processing. The effect of Post-processor on RPS system is considered neutral. There are some cases where the generation qualities are improved but also other cases where the qualities are compromised.

\begin{table*}[tb!]
    \centering
    \begin{small}
    \begin{tabular}{p{7.5cm}|p{7.5cm}} \hline
    \multicolumn{2}{l}{\textbf{Target Event}: \texttt{[BLANK]} to \textit{\textbf{maintain}} below-market rents.} \\ \hline
    RPS & RPS+Post-processor\\\hline

\cellcolor{red!15}Wilhelmina's stock \red{\textit{\textbf{dropped}}} 8 percent in early 2005 after a lull, but more actively revived the strong dollar, lending to an expected influx of funds &  Mr. Berg last year \green{\textit{\textbf{signed}}} a two-year contract with the firm\\\hline

The City Council's 10-member City Planning Committee \green{\textit{\textbf{voted}}} 11 to 6 yesterday & Another landlord is \green{\textit{\textbf{assembling}}} a plan \\\hline

\cellcolor{red!15}property values are \red{\textit{\textbf{rising}}}, rising as the vacancy rate is expected & \cellcolor{red!15}Ms. Cianci, 32, is \red{\textit{\textbf{building}}} an intercom tower \\\hline

In less than a year, such improvements have \green{\textit{\textbf{increased}}} through acquisitions and capitalization at New York City police stations, legal firms and cruise ships, suggesting that housing can be bought & \cellcolor{red!15}Her workers \red{\textit{\textbf{hit}}} a brick floor a few years ago \\\hline

\cellcolor{red!15}Matthew Hallico, president of the General Electric Company in Manhattan, and Robert Chrisin, a sales vice partner at Ira G. Albrecht His comments about the incentive package \red{\textit{\textbf{raised}}} many concerns about how it works, as well as what shareholders might do & Last week, Lloyd Pound, the influential Wall Street analyst, \green{\textit{\textbf{gave}}} 75 percent shares of his money \\\hline\hline

\multicolumn{2}{p{15cm}}{\textbf{Target Event}: By about 10 p.m. , the proposals \textit{\textbf{appeared}} dead for now \texttt{[BLANK]}.} \\ \hline
RPS & RPS+Post-processor \\\hline

after the judge, Col Richard Kultura of Thailand, \green{\textit{\textbf{signaled}}} the end of his sentence & after negotiators from both parties \green{\textit{\textbf{reconvened}}} in Davis Park to talk things out \\ \hline

\cellcolor{red!15}after city Hall \red{\textit{\textbf{learned}}} it would begin public comment on ways it could add 27,000 new jobs to the island &  after the researchers \green{\textit{\textbf{analyzed}}} DEMIC data on children and early adults whose ages began at 8 or 15\\ \hline

as they were \green{\textit{\textbf{rejected}}} by legislative leaders & \cellcolor{red!15}after Google \red{\textit{\textbf{released}}} its pie-in-pie template during an extensive public presentation \\ \hline

when Mayor Mark Meehan \green{\textit{\textbf{heeded}}} all of his smaller complaints & \cellcolor{red!15}after the Council \red{\textit{\textbf{passed}}} it on Monday\\ \hline

after the State Senate \green{\textit{\textbf{voted}}} yes on key issues &  \cellcolor{red!15}after the developer, Trivata Films of New Orleans, \red{\textit{\textbf{agreed}}} to pay up to \$14 million over seven years\\ \hline\hline

\multicolumn{2}{p{15cm}}{\textbf{Target Event}: \texttt{[BLANK]} to \textit{\textbf{scout}} potential recruits.} \\ \hline
RPS & RPS+Post-processor \\\hline
The pending replacement of Carl Crawford has \green{\textit{\textbf{enticed}}} some intelligence officials and top Qaeda leaders &
\cellcolor{red!15}About the same time, Mr. Booker \red{\textit{\textbf{elicited}}} state financing for another program that provided some of the funds through the Police Department's National Guard to help workers find mental illness or
\\\hline

Employees are \green{\textit{\textbf{giving}}} them the opportunity &
\cellcolor{red!15}In 2005 , Mr. SCAD \red{\textit{\textbf{sent}}} students from Iowa and Ohio to visit Johns Hopkins
\\\hline

Most Somalis \green{\textit{\textbf{want}}} a law that would enable them &
In championing the elite classes last week, public school teachers \green{\textit{\textbf{mounted}}} an extensive publicity campaign to persuade parents
\\\hline

Shortly after Katrina , Post servicemen were \green{\textit{\textbf{chasing}}} selectors after the storm 's onset &  \cellcolor{red!15}Joel Packer, a Detroit Pistons and assistant coach with Brigham \red{\textit{\textbf{captured}}} a larger campus and invited the scouts
\\\hline
 
\cellcolor{red!15}Ever since college opened in 1983 , he \red{\textit{\textbf{shopped}}} for school assignments & As the trend forward moves into next season , larger colleges are \green{\textit{\textbf{beginning}}} with faculty members from 75 sites on an extensive bioharker scholarship site\\ \hline
    \end{tabular}
    \end{small}
    \caption{Top 5 generation examples from RPS and RPS+Post-processor. Green colored events are considered legitimate preconditions and red colored ones are not. A red colored cell indicates invalid precondition text. As the examples show, the effect of Post-processor on RPS system is neutral -- in some cases, the generation qualities are improved but compromised in other cases.}
    \label{tab:gen_examples_rps+} \vspace{-0.15in}
\end{table*}

\subsection{Manual Evaluation}
\begin{figure*}[ht!]
    \centering
    \includegraphics[width=0.97\textwidth]{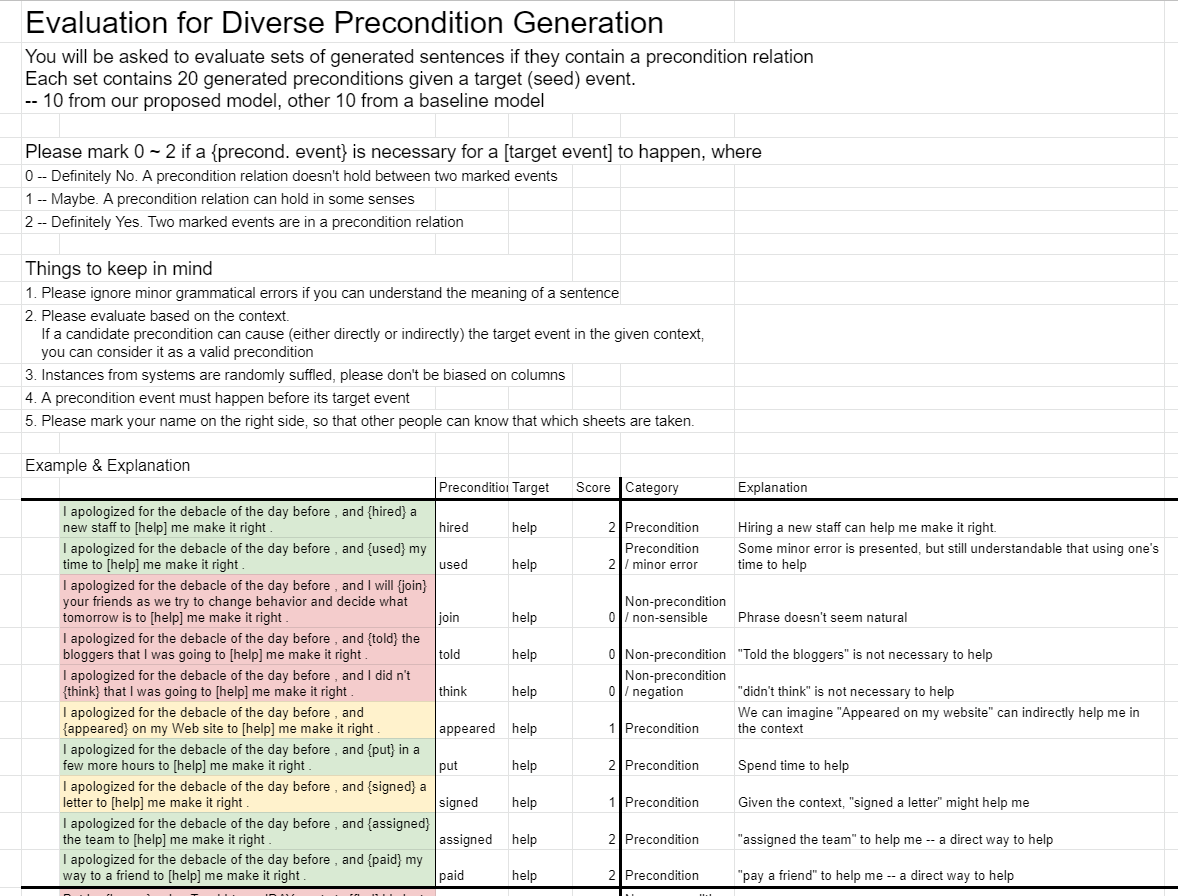} 
    \caption{Manual evaluation instruction} \vspace{-0.15in}
    \label{fig:eval_instruction}
\end{figure*}

\noindent \textbf{Evaluation Instruction} Figure \ref{fig:eval_instruction} shows the evaluation instruction that we provided to annotators.

\noindent \textbf{Evaluation rating distribution} Table \ref{tab:rating_dist} shows the distribution of voted ratings by annotators. On average, \sys\ got higher ratings than RPS and RPS got highest votes in ``Not a Precondition.''

\begin{table*}[tb!]
    \centering
    \nblue{Evaluation rating distribution}\\\vspace{.1cm}
    \begin{tabular}{l|c|c|c}
         \nblue{Model} & \nblue{Not a Precond.} & \nblue{Maybe} & \nblue{Def. a Precond.}\\\hline\hline
         RPS & 38.6\% & 27.3\% & 34.1\%\\
         \sys & 29.8\% & 30.3\% & 39.9\%\\\hline
    \end{tabular}
    \caption{Evaluation rating distribution. On average, \sys\ got higher ratings than RPS.}
    \vspace{-0.1in}
    \label{tab:rating_dist}
\end{table*}

\end{document}